# An Experimental Comparison of Naive Bayesian and Keyword-Based Anti-Spam Filtering with Personal E-mail Messages

Ion Androutsopoulos, John Koutsias, Konstantinos V. Chandrinos and Constantine D. Spyropoulos
Software and Knowledge Engineering Laboratory
Institute of Informatics and Telecommunications
National Centre for Scientific Research "Demokritos"
153 10 Ag. Paraskevi, Athens, Greece
e-mail: {ionandr, jkoutsi, kostel, costass}@iit.demokritos.gr

## Abstract

The growing problem of unsolicited bulk e-mail, also known as "spam", has generated a need for reliable anti-spam e-mail filters. Filters of this type have so far been based mostly on manually constructed keyword patterns. An alternative approach has recently been proposed, whereby a Naive Bayesian classifier is trained automatically to detect spam messages. We test this approach on a large collection of personal e-mail messages, which we make publicly available in "encrypted" form contributing towards standard benchmarks. We introduce appropriate cost-sensitive measures, investigating at the same time the effect of attribute-set size, training-corpus size, lemmatization, and stop lists, issues that have not been explored in previous experiments. Finally, the Naive Bayesian filter is compared, in terms of performance, to a filter that uses keyword patterns, and which is part of a widely used e-mail reader.

## Keywords

filtering/routing; text categorization; machine learning and IR; evaluation (general); test collections

## 1. Introduction

In recent years, the increasing popularity and low cost of e-mail have attracted the attention of direct marketers. Using readily available bulk-mailing software and large lists of e-mail addresses, typically harvested from web pages and newsgroup archives, it is now possible to send blindly unsolicited messages to thousands of recipients at essentially no cost. As a result, it is becoming increasingly common for users to receive daily large quantities of unsolicited bulk e-mail, known as *spam*, advertising anything from vacations to get-rich schemes. The term *Unsolicited Commercial E-mail* (UCE) is also used in the literature. We use "spam" with a broader meaning, that does not exclude unsolicited bulk e-mail sent for non-commercial purposes (e.g. to communicate a message from a sectarian group).

Spam messages are annoying to most users, as they waste their time and clutter their mailboxes. They also cost money to users with dial-up connections, waste bandwidth, and may expose minors to unsuitable content (e.g. when advertising pornographic sites). A 1997 study [3] found that spam messages constituted approximately 10% of the incoming messages to a corporate network. The situation seems to be worsening, and without appropriate counter-measures, spam messages could eventually undermine the usability of e-mail.

Anti-spam legal measures are gradually being adopted, but they have had a very limited effect so far.[1] Of more direct value are *anti-spam filters*, software tools that attempt to block automatically spam messages.[2] Apart from blacklists of frequent spammers and lists of trusted users, which can be incorporated into any anti-spam strategy, these filters have so far relied mostly on manually constructed keyword patterns (e.g. blocking messages whose bodies contain "be over 21"). To be most effective and to avoid the risk of accidentally deleting non-spam messages, hereafter called *legitimate messages*, these patterns need to be manually tuned to the incoming e-mail of each user. Fine-tuning the patterns, however, requires time and expertise that are not always available. Even worse, the characteristics of spam messages (e.g. topics, frequent terms) change over time, requiring the keyword patterns to be updated periodically [3]. It is, therefore, desirable to develop anti-spam filters that will learn *automatically* how to block spam messages by processing previously received spam and legitimate messages.

Sahami et al. [23] recently proposed using a machine learning algorithm to construct a filter of this type. They trained a Naive Bayesian classifier [7] [20] on manually categorized legitimate and spam messages, reporting impressive performance on unseen messages. Although machine learning algorithms have been applied to several text categorization tasks (e.g. [1], [5], [17], [18], [19]), including applications where the goal is to thread e-mail [18], to classify e-mail into folders [2] [21], or to identify interesting news articles ([14]; see also [11]), to the best of our knowledge Sahami et al.'s experiments constitute the only previous attempt to apply machine learning to anti-spam filtering.

It may come as a surprise that text categorization techniques can be effective in anti-spam filtering. Unlike most text categorization tasks, it is the *act* of blindly mass-mailing an unsolicited message that makes it spam, not its actual *content*:

---

[1] Consult http://www.cauce.org , http://spam.abuse.net , and http://www.junkemail.org .
[2] See, for example, http://www.tucows.com .

any otherwise legitimate message becomes spam if blindly mass-mailed. Nevertheless, it seems that the language of current spam messages constitutes a distinctive genre, and that the topics of most current spam messages are rarely mentioned in legitimate messages, making it possible to train successfully a text classifier for anti-spam filtering.

Past experience in text categorization has proven the beneficial role of publicly available manually categorized document collections, like the Reuters corpus [16], that can be used as benchmarks to compare alternative techniques. For the purposes of anti-spam filtering, such benchmark collections would ideally consist of manually categorized legitimate and spam messages as received by real users. Making a collection of this sort, however, publicly available in raw form would violate the privacy of the recipients and senders of legitimate messages. We show, however, that it is possible to "encrypt" the collection's messages in a way that respects privacy issues, while still leading to useful public collections.[3]

We test Sahami et al.'s approach on an encrypted collection of legitimate and spam messages, dubbed *PU1 corpus*, which we make publicly available contributing towards standard benchmarks.[4] Unlike previous experiments, we employ ten-fold cross-validation, which makes our results less prone to random variation. Our investigation also examines the effect of attribute-set size, training-corpus size, lemmatization, and stop-lists, issues that were not explored in Sahami et al.'s experiments. Legitimate messages that can be easily identified using lists of trusted users have not been included in PU1, leading to a "tougher" corpus than the one used by Sahami et al.; nevertheless, our results confirm Sahami et al.'s high precision and recall figures. We argue, however, that anti-spam filters must be evaluated using measures that incorporate a decision-theoretic notion of cost; and a cost-sensitive evaluation reveals that additional safety nets are necessary for anti-spam filtering to be viable in practice. Further evaluation shows that the Naive Bayesian filter is by far superior to a keyword-based anti-spam filter that is included in a widely used e-mail reader.

The remaining of this paper is organized as follows: section 2 introduces the Naive Bayesian Classifier and Sahami et al.'s results; section 3 discusses how the PU1 corpus was assembled, and presents the results we obtained with the Naive Bayesian classifier and the keyword-based filter; section 4 introduces cost-sensitive evaluation measures; and section 5 concludes.

## 2. The Naive Bayesian classifier

The Naive Bayesian classifier [7] [20] assumes that each document (in our case, each message) is represented by a vector $\vec{x} = \langle x_1, x_2, x_3, \ldots, x_n \rangle$, where $x_1, \ldots, x_n$ are the values of attributes $X_1, \ldots, X_n$, much as in the vector space model [24].

Following Sahami et al., we use binary attributes, i.e. $X_i = 1$ if the message has the property represented by $X_i$, and $X_i = 0$ otherwise. Our experiments were conducted using only word-attributes, i.e. each attribute shows whether or not a particular word (eg. "adult") is present in the message. It is possible, however, to use additional attributes corresponding to phrases (e.g. "be over 21") or non-textual properties (e.g. whether or not the message contains attachments); we return to this point below.

Following Sahami et al., we use *mutual information* ( $MI$ ) to select among all possible attributes (in our case, all possible word-attributes). We compute the mutual information $MI(X;C)$ of each candidate attribute $X$ with the category-denoting variable $C$ as:

$$\sum_{x \in \{0,1\}, c \in \{spam, legitimate\}} P(X = x, C = c) \cdot \log \frac{P(X = x, C = c)}{P(X = x) \cdot P(C = c)}$$

We then select the attributes with the highest mutual information values. The probabilities $P(X,C)$, $P(X)$, and $P(C)$ are estimated from a training corpus as frequency ratios.[5]

From Bayes' theorem and the theorem of total probability, it follows that the probability that a document with vector $\vec{x} = \langle x_1, \ldots, x_n \rangle$ belongs to category $c$ is:

$$P(C = c \mid \vec{X} = \vec{x}) = \frac{P(C = c) \cdot P(\vec{X} = \vec{x} \mid C = c)}{\sum_{k \in \{spam, legitimate\}} P(C = k) \cdot P(\vec{X} = \vec{x} \mid C = k)}$$

In practice, the probabilities $P(\vec{X} \mid C)$ are impossible to estimate without simplifying assumptions, because the possible values of $\vec{X}$ are too many and there are also data sparseness problems. The Naive Bayesian classifier assumes that $X_1, \ldots, X_n$ are conditionally independent given the category $C$, which allows us to compute $P(C = c \mid \vec{X} = \vec{x})$ as:

$$\frac{P(C = c) \cdot \prod_{i=1}^{n} P(X_i = x_i \mid C = c)}{\sum_{k \in \{spam, legitimate\}} P(C = k) \cdot \prod_{i=1}^{n} P(X_i = x_i \mid C = k)}$$

$P(X_i \mid C)$ and $P(C)$ are easy to estimate from the frequencies of the training corpus. A large number of empirical studies have found the Naive Bayesian classifier to be surprisingly effective [6] [15], despite the fact that the assumption that $X_1, \ldots, X_n$ are conditionally independent is usually (including our word-attributes case) overly simplistic.[6]

---

[3] Another approach is to use a collection consisting of spam messages, and legitimate messages extracted from publicly available archives of e-mail lists. We explored this direction in previous work to be reported elsewhere.
[4] PU1 is available from the publications section of http://www.iit.demokritos.gr/~ionandr .

[5] Consult [20] for more elaborate estimates that we hope to investigate in future work.
[6] Consult [10] for forms of Bayesian classifiers with less restrictive independence assumptions.

| Attributes | Total Messages | Testing Messages | Spam Messages (%) | Spam Precision (%) | Spam Recall (%) |
|---|---|---|---|---|---|
| words only | 1789 | 251 | 88.2 | 97.1 | 94.3 |
| words + phrases | 1789 | 251 | 88.2 | 97.6 | 94.3 |
| words + phrases + non-textual | 1789 | 251 | 88.2 | 100.0 | 98.3 |
| words + phrases + non-textual | 2815 | 222 | ~20 | 92.3 | 80.0 |

Table 1: Results of Sahami et al. (500 attributes, $t = 0.999$, $\lambda = 999$)

In anti-spam filtering, mistakenly blocking a legitimate message (classifying a legitimate message as spam) is generally more severe an error than letting a spam message pass the filter (classifying a spam message as legitimate). We use $L \rightarrow S$ (legitimate to spam) and $S \rightarrow L$ (spam to legitimate) to denote the two error types, respectively, and invoking a decision-theoretic notion of cost, we assume that $L \rightarrow S$ is $\lambda$ times more costly than $S \rightarrow L$. We classify a message as spam if the following *classification criterion* is met:

$$\frac{P(C = spam | \vec{X} = \vec{x})}{P(C = legitimate | \vec{X} = \vec{x})} > \lambda$$

To the extent that the independence assumption holds and the probability estimates are accurate, a classifier based on this criterion achieves optimal results [7]. In our case, $P(C = spam | \vec{X} = \vec{x}) = 1 - P(C = legitimate | \vec{X} = \vec{x})$, and the criterion above is equivalent to:

$$P(C = spam | \vec{X} = \vec{x}) > t \text{, with } t = \frac{\lambda}{1+\lambda}, \lambda = \frac{t}{1-t}$$

In the experiments of Sahami et al., the threshold $t$ was set to 0.999, which corresponds to $\lambda = 999$. This means that mistakenly blocking a legitimate message was taken to be as bad as letting 999 spam messages pass the filter. When blocked messages are discarded without further processing, setting $\lambda$ to such a high value is reasonable, because in that case most users would consider losing a legitimate message unacceptable. Configurations with additional safety nets are possible, however, and lower $\lambda$ values would be reasonable in those cases.

For example, rather than being deleted, a blocked message could be returned to the sender, along with an automatically inserted apology paragraph. The extra paragraph would explain that the message was blocked by an anti-spam filter, and it would ask the sender to forward the message to a different, private un-filtered e-mail address of the recipient (see also [12]). The private address would never be advertised (e.g. on web pages or newsgroups), making it unlikely to receive spam mail directly. The apology paragraph could also include a frequently changing riddle (e.g. "Include in the subject the capital of France.") to ensure that spam messages are not forwarded automatically to the private address by robots that scan returned messages for new e-mail addresses. Messages sent to the private address without the correct riddle answer would be deleted automatically. (Spammers cannot afford the time to answer by hand thousands of riddles.)

In the scenario of the previous paragraph, $\lambda = 9$ ($t = 0.9$) seems more reasonable: blocking a legitimate message is penalized mildly more than letting a spam message pass, to account for the fact that recovering from a blocked legitimate message requires overall more work (counting the sender's extra work to repost it) than recovering from a spam message that passed the filter (deleting it manually). If the recipient does not care about the extra work imposed on the sender, then even $\lambda = 1$ ($t = 0.5$) can be acceptable.

Table 1 shows the results that Sahami et al. reported, using $\lambda = 999$ ($t = 0.999$). (We omit an experiment on detecting spam subcategories, which did not show promising results.) Assuming that $n_{L \rightarrow S}$ and $n_{S \rightarrow L}$ are the numbers of $L \rightarrow S$ and $S \rightarrow L$ errors, and that $n_{L \rightarrow L}$ and $n_{S \rightarrow S}$ count the correctly treated legitimate and spam messages respectively, spam recall ($SR$) and spam precision ($SP$) are defined as follows:

$$SR = \frac{n_{S \rightarrow S}}{n_{S \rightarrow S} + n_{S \rightarrow L}} \qquad SP = \frac{n_{S \rightarrow S}}{n_{S \rightarrow S} + n_{L \rightarrow S}}$$

The first three experiments of table 1 were performed with a collection of 1789 messages, consisting of 211 legitimate messages that users had saved and 1578 spam messages. In the first experiment only word-attributes were used. In the second experiment, phrasal candidate attributes were added (e.g. corresponding to the phrases "be over 21", "only $"). In the third experiment, additional non-textual candidate attributes were inserted (e.g. whether or not the message contains attachments, or a high proportion of non alphanumeric characters). The phrasal and non-textual candidate attributes were constructed by hand. All candidate attributes (word, phrasal, non-textual) were subjected to a single $MI$-based selection. The fourth experiment was similar to the third one, but it was performed with a different collection, containing all the messages a user had received over an entire year. The 500 attributes with the highest $MI$ were used in all cases, but no rigorous testing was conducted to select this number of attributes.

While phrasal and non-textual attributes were found to improve results, they introduce a manual configuration phase, since one has to select manually phrases and non-textual properties to be treated as candidate attributes. Our target was to explore fully automatic anti-spam filtering, and hence we limited ourselves to word-attributes. We hope to incorporate in future work automatic techniques, similar to those used in term extraction [9], to locate candidate phrasal attributes automatically.

## 3. Experiments with the PU1 corpus

We now turn to the corpus that we used, PU1, and our experiments. The PU1 corpus consists of 1099 messages:

481 spam messages. These are all the spam messages that the first author received over a period of 22 months, excluding non-English messages (so far, these are very rare). Duplicates of spam messages sent on the same day were also not included (these are common, but they are very easy to detect with conventional technology).

618 legitimate messages. These were derived as follows. First, all the English legitimate messages that the first author had received and saved (excluding self-addressed messages) over a period of 36 months were collected (1182 messages). Many of the collected messages were from people with which the first author has (or had) regular correspondence, mostly colleagues and friends, that are unlikely to send spam messages. To ensure that the anti-spam filter never misclassifies messages from those senders, the filter's users can be instructed to insert into their address books people they find they correspond regularly with (this is very easy in modern e-mail readers). Messages received from senders in a user's address book can then be classified blindly as legitimate, without applying the anti-spam filter on them. To emulate this mechanism, we deleted from the 1182 messages all but the earliest five messages of each sender (leaving 618). We assume that by the time the sixth legitimate message of a particular sender arrives, the sender will have been inserted into the address book, and the anti-spam filter will not have to examine messages from this sender any more.

For our experiments, we implemented the Naive Bayesian classifier on the GATE platform [4]. Our implementation includes an English lemmatizer that converts each word to its base form (e.g. "earned" becomes "earn"), and a stop-list module that removes from each message the 100 most frequent words of the British National Corpus (BNC).[7] These two modules can be enabled or disabled to measure their effect.

Attachments and HTML tags were removed from all messages. To respect privacy issues, in the publicly available version of PU1, fields other than "Subject:" were removed, and each token (word, number, punctuation symbol, etc.) in the bodies or subjects of the messages was replaced by a unique number, the same number throughout all the messages. For example:

> *From: spammer@spamcompany.com*
> *To: spamtarget@provider.com*
> *Subject: Get rich now !*
>
> *Click here to get rich ! Try it now !*

becomes:

> *Subject: 1 2 3 4*
>
> *5 6 7 1 2 4 8 9 3 4*

There are actually four "encrypted" versions of the publicly available PU1 corpus, one for each combination of enabled/disabled stop-list and lemmatizer. The correspondence between tokens and numbers is not released, making it very difficult to figure out what the messages say.[8] This encryption scheme places some limitations on the resources that can be exploited when experimenting with PU1. For example, thesauri or part-of-speech taggers cannot be used, since the exact words of the messages are unknown. PU1, however, can still be used to experiment with any classification technique that relies only on frequency and co-occurrence statistics (the Naive Bayesian classifier and most of the machine learning techniques cited in section 1 fall into this category).

In all our experiments with the Naive Bayesian classifier, we used *ten-fold cross validation* to reduce random variation. That is, PU1 was partitioned randomly into ten parts, and each experiment was repeated ten times, each time reserving a different part for testing, and using the remaining nine parts for training. Results were then averaged over the ten runs. In our first series of experiments, we varied the number of retained attributes (attributes with highest $MI$) from 50 to 700 by a step of 50, for all four combinations of enabled/disabled lemmatizer and stop-list, and for three thresholds: $t = 0.999$ ($\lambda = 999$), $t = 0.9$ ($\lambda = 9$), and $t = 0.5$ ($\lambda = 1$). As discussed in section 2, these thresholds are taken to represent three usage scenarios of the anti-spam filter: deleting blocked messages immediately; asking the sender to re-post them to a private address and accounting for the sender's extra work; and asking the sender to re-post them to a private address ignoring the sender's extra work. Figures 1 – 3 show the spam recall and spam precision that the Naive Bayesian classifier achieved. There are different points for different numbers of retained attributes.

We also implemented a simpler filter that uses the (presumably, manually constructed) keyword-based patterns of the anti-spam filter of Microsoft Outlook 2000.[9] These are 58 patterns, looking for particular keywords in the body or header fields of the messages (e.g. "Body contains ',000' AND Body contains '!!' AND Body contains '$'"). The keyword-based filter was applied on an unencrypted version of the PU1 corpus, achieving spam precision 95.15% and spam recall 53.01%. (These results were obtained using a case-insensitive version of the patterns. The original case-sensitive patterns achieved spam precision 95.45% and spam recall 39.29%.)

---

[7] GATE, including the lemmatizer we used, is available from http://www.dcs.shef.ac.uk/research/groups/nlp/gate. BNC frequencies are available from ftp://ftp.itri.bton.ac.uk/pub/bnc. We also experimented with stop-lists containing frequent words of particular parts of speech (e.g. verbs only), but results showed no significant difference.

[8] Decryption experts may be able to recover chunks of the original text based on frequency and co-occurrence patterns. Given the uninteresting nature of the messages in PU1, however, it is difficult to imagine why one would want to waste time on this exercise.

[9] "Microsoft Outlook" is a trademark of Microsoft Corporation. Outlook's on-line documentation points to a file that contains the patterns of its anti-spam filter.

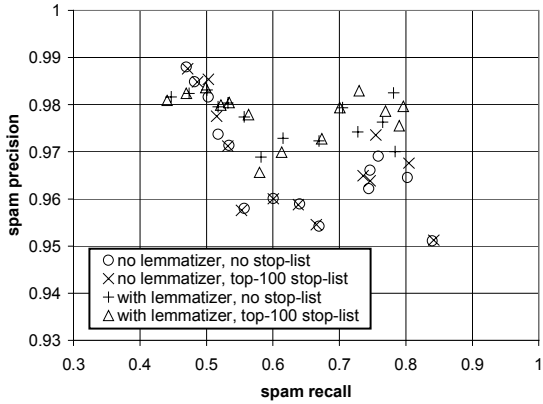

Figure 1: Spam precision and recall of the Naive Bayesian filter for $t = 0.5$ ($\lambda = 1$)

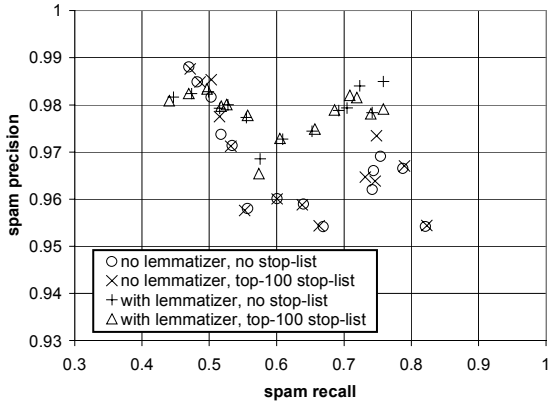

Figure 2: Spam precision and recall of the Naive Bayesian filter for $t = 0.9$ ($\lambda = 9$)

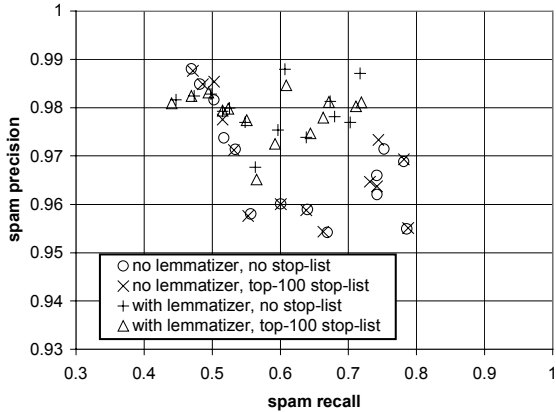

Figure 3: Spam precision and recall of the Naive Bayesian filter for $t = 0.999$ ($\lambda = 999$)

Figures 1 – 3 show a gradual increase of spam precision and decrease of spam recall as $\lambda$ increases. This is due to the fact that for higher $\lambda$ values the classifier needs to be more "certain" before blocking a message as spam, which increases its precision but reduces the number of spam messages it blocks. Overall, the Naive Bayesian filter achieves impressive spam recall and precision combinations at all three thresholds, verifying similar findings by Sahami et al. The Naive Bayesian filter also outperforms the keyword-based filter at most numbers of retained attributes (most of the points in figures 1 – 3 are over 95.15% precision and 53.01% recall). The spread of the points in the figures, however, shows that the number of retained attributes has an important influence on spam recall and precision, and hence it cannot be chosen without careful experimentation. Without a single evaluation measure, that would be used instead of spam recall and precision, it is difficult to decide which number of attributes (or equivalently which recall-precision combination) leads to the best results, and whether or not the lemmatizer and stop-list have statistically significant effects. The single evaluation measure must also be sensitive to the fact that the $L \rightarrow S$ and $S \rightarrow L$ errors are not assigned equal costs.[10] We introduce appropriate cost-sensitive evaluation measures in the next section.

## 4. Cost-sensitive evaluation

In classification tasks, performance is often measured in terms of *accuracy* ($Acc$) or *error rate* ($Err = 1 - Acc$). If $N_L$ and $N_S$ are the total numbers of legitimate and spam messages to be classified by the filter, respectively, and $n_{L \rightarrow L}$, $n_{S \rightarrow S}$, $n_{L \rightarrow S}$, $n_{S \rightarrow L}$ are as in section 2, then:

$$Acc = \frac{n_{L \rightarrow L} + n_{S \rightarrow S}}{N_L + N_S} \qquad Err = \frac{n_{L \rightarrow S} + n_{S \rightarrow L}}{N_L + N_S}$$

Accuracy and error rate assign equal weights to the two error types ($L \rightarrow S$ and $S \rightarrow L$). When formulating the classification criterion (section 2), however, we assumed that $L \rightarrow S$ is $\lambda$ times more costly than $S \rightarrow L$. To make accuracy and error rate sensitive to this cost difference, we treat each legitimate message as if it were $\lambda$ messages. That is, when a legitimate message is blocked, this counts as $\lambda$ errors; and when it passes the filter, this counts as $\lambda$ successes. This leads to the following definitions of *weighted accuracy* ($WAcc$) and *weighted error rate* ($WErr = 1 - WAcc$):

$$WAcc = \frac{\lambda \cdot n_{L \rightarrow L} + n_{S \rightarrow S}}{\lambda \cdot N_L + N_S} \qquad WErr = \frac{\lambda \cdot n_{L \rightarrow S} + n_{S \rightarrow L}}{\lambda \cdot N_L + N_S}$$

The values of accuracy and error rate (or our weighted versions of them) are often misleadingly high. To get a clear picture of the performance of a classifier, it is common to compare its accuracy or error rate to those of a simplistic "baseline" approach. We take the case where no filter is present to be our baseline: legitimate messages are (correctly) never blocked, and spam messages (mistakenly) always pass the filter. The weighted accuracy and weighted error rate of the baseline are:

---

[10] The F-measure, often used in information retrieval and extraction to combine recall and precision (e.g. [22]) cannot be used here, because its weighting factor cannot be easily related to our notion of cost.

| Filter used ("NB" is Naive Bayesian) | λ | no. of attr. | spam recall (%) | spam precision (%) | weighted accuracy (%) | TCR |
|---|---|---|---|---|---|---|
| (a) NB bare | 1 | 50 | 83.98 | 95.11 | 91.076 | 4.90 |
| (b) NB with stop-list | 1 | 50 | 84.19 | 96.76 | 91.167 | 4.95 |
| (c) NB with lemmatizer | 1 | 100 | 78.14 | 98.25 | 89.796 | 4.29 |
| (d) NB with lemmatizer and stop-list | 1 | 100 | 79.60 | 97.96 | 90.341 | 4.53 |
| *Keyword patterns (case insensitive)* | 1 | – | 53.01 | 95.15 | 78.253 | 2.01 |
| *Baseline (no filter)* | 1 | – | 0 | ∞ | 56.233 | 1 |
| (e) NB bare | 9 | 100 | 78.77 | 96.65 | 96.378 | 2.20 |
| (f) NB with stop-list | 9 | 150 | 74.83 | 97.34 | 96.508 | 2.28 |
| (g) NB with lemmatizer | 9 | 100 | 75.86 | 98.50 | 97.183 | 2.83 |
| (h) NB with lemmatizer and stop-list | 9 | 100 | 75.86 | 97.91 | 96.886 | 2.56 |
| *Keyword patterns (case insensitive)* | 9 | – | 53.01 | 95.15 | 94.324 | 1.40 |
| *Baseline (no filter)* | 9 | – | 0 | ∞ | 92.040 | 1 |
| (i) NB bare | 999 | 700 | 46.96 | 98.80 | 99.475 | 0.15 |
| (j) NB with stop-list | 999 | 700 | 47.17 | 98.76 | 99.475 | 0.15 |
| (k) NB with lemmatizer | 999 | 50 | 60.68 | 98.79 | 99.322 | 0.11 |
| (l) NB with lemmatizer and stop-list | 999 | 600 | 49.45 | 98.31 | 99.313 | 0.11 |
| *Keyword patterns (case insensitive)* | 999 | – | 53.01 | 95.15 | 97.862 | 0.04 |
| *Baseline (no filter)* | 999 | – | 0 | ∞ | 99.922 | 1 |

Table 2: Results on the PU1 corpus using the best number of attributes (1099 total messages, 43.8% spam, 10-fold cross validation, number of attributes ranging from 50 to 700 by 50)

$$WAcc^b = \frac{\lambda \cdot N_L}{\lambda \cdot N_L + N_S} \quad WErr^b = \frac{N_S}{\lambda \cdot N_L + N_S}$$

We also introduce the *total cost ratio* (*TCR*), defined below, which allows the performance of an anti-spam filter to be compared easily to that of the baseline:

$$TCR = \frac{WErr^b}{WErr} = \frac{N_S}{\lambda \cdot n_{L \to S} + n_{S \to L}}$$

Greater *TCR* values indicate better performance. For $TCR < 1$, the baseline (not using the filter) is better. If cost is proportional to wasted time, an intuitive meaning for *TCR* is the following: it measures how much time is wasted to delete manually all spam messages when no filter is used ($N_S$), compared to the time wasted to delete manually any spam messages that passed the filter ($n_{S \to L}$) plus the time needed to recover from mistakenly blocked legitimate messages ($\lambda \cdot n_{L \to S}$).

Table 2 summarizes the results we obtained with the Naive Bayesian filter on PU1 for $\lambda = 1, 9, 999$, with the lemmatizer and stop-list enabled or disabled, and using the best numbers of attributes (those that lead to the highest *TCR* scores in each case). It also shows the corresponding results of the keyword-based filter and the baseline.[11] Figures 4 – 6 show the *TCR* scores for the three $\lambda$ values at different numbers of attributes. In all the experiments with the Naive Bayesian (NB) filter, ten-fold cross validation was used, and average *WAcc* is reported.

In ten-fold cross validation experiments, *TCR* is computed as $WErr^b$ divided by the average *WErr*.[12]

For $\lambda = 1$, the Naive Bayesian filter achieves $TCR > 1$ at all numbers of attributes (figure 4). The keyword-based filter also scores $TCR > 1$, but the Naive Bayesian filter is better up to 550 attributes. Increasing the number of attributes of the Naive Bayesian filter beyond a certain point degrades its performance, because attributes with low *MI* do not discriminate well between spam and legitimate messages. Paired single-tailed t-tests on *WAcc* confirm at $p < 0.001$ that configurations (a), (b), (c), and (d) of the Naive Bayesian filter (table 2) are all significantly better than the keyword-based filter and the baseline. Enabling the lemmatizer or the stop-list does not seem to have any significant effect. Indeed, paired single-tailed t-tests at $p < 0.05$ do not confirm the hypotheses of table 2 about (a), (b), (c), (d) themselves (e.g. that (b) is better than (c)); and at $p < 0.1$, the only one of these hypotheses that is confirmed is that (d) is better than (c).

For $\lambda = 9$, again both filters achieve constantly $TCR > 1$, with the Naive Bayesian being always better (figure 5). The stop-list has hardly any effect, while the lemmatizer seems to lead to a noticeable improvement. The improvement of the lemmatizer, however, is not significant from a statistical point of view: the t-tests confirm the hypothesis of table 2 that (g) is better than (e) and (f) (and also (h)) only at $p < 0.1$, not $p < 0.05$. No other hypothesis of table 2 about (e), (f), (g), (h) themselves (e.g. that (h) is better than (e)) is confirmed at $p < 0.1$. The t-tests, however, confirm at $p < 0.001$ that all the configurations

---

[11] The TCR scores with case sensitive keyword-patterns were 1.60, 1.29, and 0.05, respectively.

[12] It is important *not* to compute the overall TCR as the average of the TCRs of the individual folds (repetitions), as this effectively ignores folds with TCR << 1.

(e), (f), (g), and (h) of the Naive Bayesian filter are better than the keyword-based filter and the baseline.

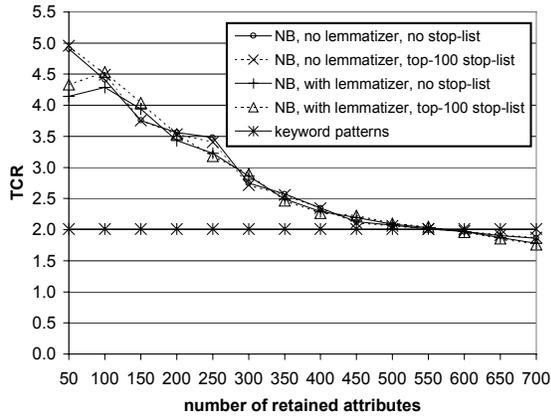

Figure 4: $TCR$ of the filters for $t = 0.5$ ($\lambda = 1$)

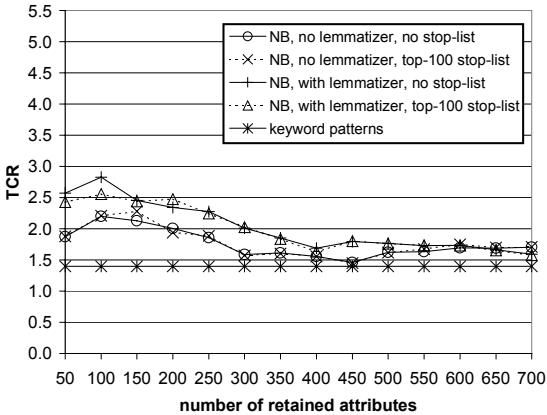

Figure 5: $TCR$ of the filters for $t = 0.9$ ($\lambda = 9$)

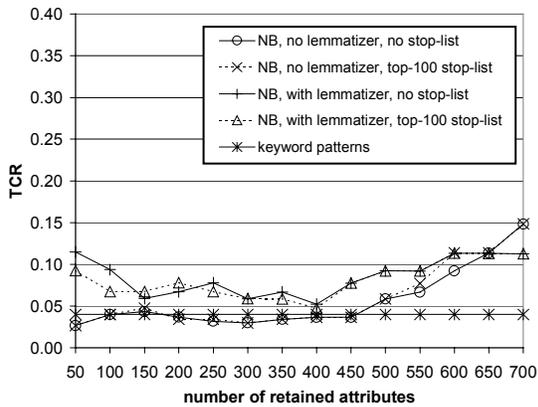

Figure 6: $TCR$ of the filters for $t = 0.999$ ($\lambda = 999$)

For $\lambda = 999$, the t-tests again confirm at $p < 0.001$ that configurations (i), (j), (k), and (l) of the Naive Bayesian filter (table 2) are all significantly better than the keyword-based filter, though none of the other hypotheses of table 2 is confirmed at $p < 0.1$. Both filters, however, score constantly $TCR < 1$ (figure 6; notice the different scale from figures 4 and 5). This is due to the very high penalty on the $L \rightarrow S$ errors, and the fact that none of the two filters manages to eliminate these errors completely. We conclude that for $\lambda = 999$, not using any of the two filters is actually better, despite their high spam precision and recall.

A remaining question is how much training data the Naive Bayesian filter needs to be effective for $\lambda = 1$ and $\lambda = 9$. To answer this question, we conducted a second series of experiments with the Naive Bayesian filter, this time varying the size of the training corpus. The corpus was again divided into ten parts, and one part was reserved for testing at every run. From each one of the remaining nine parts, only $x$ % was used for training, with $x$ ranging from 10 to 100 by 10. The experiments were performed with the best configurations of table 2, i.e. (b) and (g). Figure 7 shows the resulting $TCR$ scores for $\lambda = 1$ and $\lambda = 9$. It also shows the $TCR$ score that the keyword-based filter achieved for $\lambda = 1$ and $\lambda = 9$ on the entire corpus. It can be seen that the Naive Bayesian (NB) filter achieves $TCR > 1$ and outperforms the keyword-based filter even with very small training corpora.

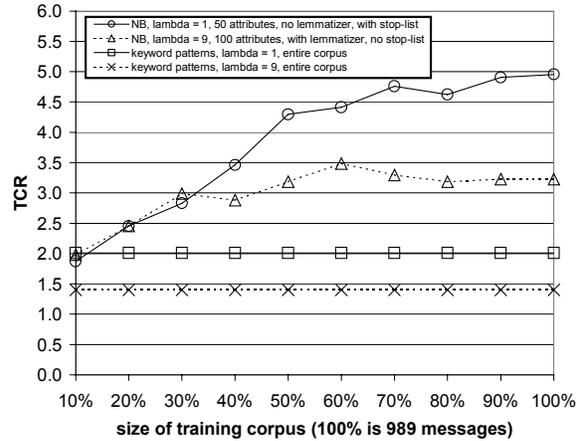

Figure 7: $TCR$ for variable size of training corpus

## 5. Conclusions

Our cost-sensitive evaluation suggests that neither the Naive Bayesian nor the keyword-based filter perform well enough to be used when $\lambda = 999$, a $\lambda$ value we employed to model the scenario where blocked messages are deleted without further processing. The performance of both filters, however, is satisfactory for $\lambda = 1$ and $\lambda = 9$. We used the latter values to model the case where a mechanism is available to re-post blocked messages to private addresses.

The Naive Bayesian filter outperforms by far the keyword-based filter, even with very small training corpora. For $\lambda = 9$ and $\lambda = 999$, its performance showed signs of improvement when a lemmatizer was added, but the improvements were not statistically significant. Adding a stop-list does not seem to have any noticeable effect, presumably because the $MI$-based attribute selection rarely picks words that are so common as those of the stop-list.

We interpret these results as implying that, with additional safety nets, automatically trainable anti-spam filters are practically viable and can have a significant effect. We plan to implement alternative anti-spam filters, based on other machine learning algorithms, and evaluate them on publicly available benchmark corpora. We expect anti-spam filtering to become an important member of an emerging family of junk-filtering tools for the Internet, which will include – among others – tools to remove unwanted advertisements [13], and block hostile or pornographic material [8] [25].

**Acknowledgments**

The authors wish to thank George Paliouras for comments on an earlier version of this paper, and Stavros Perantonis for his help with statistical significance tests. The authors are also grateful to Antonis Argyros and Lena Gaga, who contributed a collection of spam messages that was used in earlier experiments.